\documentclass[10pt,journal,compsoc]{IEEEtran}

\usepackage{amsmath}
\usepackage{amssymb}
\usepackage{algorithmic}
\usepackage{array}
\usepackage{url}
\usepackage{graphicx}
\usepackage{xcolor}
\usepackage{soul}

\begin{document}

\title{
Modeling Feature Representations for Affective Speech using Generative Adversarial Networks
}

\author{Saurabh~Sahu,~\IEEEmembership{Member,~IEEE,}
        Rahul~Gupta,~\IEEEmembership{Member,~IEEE,}
        and~Carol~Espy-Wilson,~\IEEEmembership{Senior~Member,~IEEE}
\IEEEcompsocitemizethanks{\IEEEcompsocthanksitem S. Sahu and C. Espy-Wilson are with the Department of Electrical and Computer Engineering, University of Maryland, College Park,
MD, 20740. e-mail: ssahu89@umd.edu; espy@umd.edu\protect\\

\IEEEcompsocthanksitem R. Gupta is with Amazon.com. email: gupra@amazon.com}
\thanks{Manuscript received April 19, 2005; revised August 26, 2015.}}

\markboth{Journal of \LaTeX\ Class Files,~Vol.~14, No.~8, August~2015}%
{Shell \MakeLowercase{\textit{et al.}}: Bare Demo of IEEEtran.cls for Computer Society Journals}

\IEEEtitleabstractindextext{%
\begin{abstract}
{

Emotion recognition is a classic field of research with a typical setup extracting features and feeding them through a classifier for prediction.
On the other hand, generative models jointly capture the distributional relationship between emotions and the feature profiles.
Relatively recently, Generative Adversarial Networks (GANs) have surfaced as a new class of generative models and have shown considerable success in modeling distributions in the fields of computer vision and natural language understanding. 
In this work, we experiment with variants of GAN architectures to generate feature vectors corresponding to an emotion in two ways: (i) A generator is trained with samples from a mixture prior. Each mixture component corresponds to an emotional class and can be sampled to generate features from the corresponding emotion.
(ii) A one-hot vector corresponding to an emotion can be explicitly used to generate the features.
We perform analysis on such models and also propose different metrics used to measure the performance of the GAN models in their ability to generate realistic synthetic samples. 
Apart from evaluation on a given dataset of interest, we perform  a cross-corpus study where we study the utility of the synthetic samples as additional training data in low resource conditions.

}
\end{abstract}

}

\maketitle

\IEEEdisplaynontitleabstractindextext

\IEEEpeerreviewmaketitle

\section{Introduction}
Recognizing emotions from speech has usefulness in many areas such as psychology, medicine and designing human-computer interaction systems \cite{el2011survey}. The fact that speech is easy to collect unlike physiological signals has made it a popular candidate to build models for such tasks. Typically, designing a classification system entails extracting feature vectors $\mathbf{x} \in \mathbb{R}^d$ from speech signal which could carry information about the emotional state of the speaker. A classifier is then trained to estimate the conditional probability $p(y|\mathbf{x})$ ($y \in$ set of categorical emotional labels) using the ground truth annotations. Features vectors are usually high dimensional which leads to the joint distribution $p(\mathbf{x},y)$ to lie in complex manifolds. Understanding their distribution could be the key to build robust classifiers. 
In the past, researchers have used the generative ability of models for tasks such as building emotion classification models \cite{chandrakala2009combination} . In this paper, we focus on analyzing one specific category of generative models: Generative Adversarial Networks (GANs) when applied to speech based emotion recognition. Using multiple GAN variants, we present insights on model training as well as an analysis on the quality of feature vectors generated. We also propose multiple metrics to evaluate the quality of generated samples and discuss their interpretations. With these contributions, we aim to advance the application of GANs to speech emotion recognition for tasks such as training emotion classification and synthetic feature generation.

Deep generative models such as Generative Adversarial Nets (GANs) \cite{goodfellow2014generative} and variational auto-encoders (VAEs) \cite{kingma2013auto} are popular variants of generative model attributed to their ability to capture the complexities of real world data distribution and generate realistic examples from those distributions. GANs have been successful in tasks such as image generation \cite{radford2015unsupervised}, style transfer \cite{zhang2017style} and speech enhancement \cite{pascual2017segan}. The objective of GAN training is to obtain a generator which when fed samples $\mathbf{z}$ from a lower dimensional simple distribution $p_{\mathbf{z}}$ can generate higher dimensional realistic looking data points. VAEs are probabilistic graphical models implemented using deep networks to learn an efficient lower dimensional encoding of higher dimensional feature vectors. Synthetic feature vectors can then be generated by passing the samples generated from the lower dimensional space through the VAE decoder. VAEs can be used for data generation as well as matching the lower dimensional latent space to a desired distribution, typically done by minimizing distance of the latent space with a pre-defined prior distribution such as a Gaussian distribution. 
Applications of VAE include blurry image generation \cite{arjovsky2017wasserstein}, anomaly detection \cite{an2015variational} and text generation \cite{semeniuta2017hybrid}. 
Makhzani et. al \cite{makhzani2015adversarial} propose using GAN based adversarial losses to match the distribution of latent space to the prior distribution. This further allowed them to match the latent space with more complex distributions than a Gaussian distribution.  In \cite{sahu2018adversarial}, we enforce the latent space to resemble a mixture of four Gaussians, each mixture component spanning the latent codes obtained from samples belonging to one of the four emotion classes : angry, sad, neutral and happy. Synthetic samples belonging to a particular class can then be generated by sampling from the corresponding mixture component and passing it through the decoder of a trained adversarial auto-encoder. Note that we imposed the condition that the generated data should lie in four clusters by choosing a prior which has four mixture components. Another way to enforce the clustering would be to feed the generator with an additional label vector along with points sampled from the prior and then maximize the mutual dependence between the generated synthetic samples and the input label vectors as done in an infoGAN \cite{chen2016infogan} framework. We borrow ideas from these GAN and auto-encoder based frameworks to build generative models that can synthesize feature vectors when fed with samples from the prior distribution.

In Section~\ref{sec:related}, we look at previous attempts by researchers to benefit from generative models in building emotion classification models. In Section~\ref{sec:background} we give a brief overview of the GAN frameworks that has been used by the vision community for synthetic image generation. Section~\ref{sec:exp} explains our experimental setup. We talk about the datasets used in our experiments. We describe the auto-encoder and GAN based architectures and their training methodology for synthetic data generation. We also explain the proposed metrics. In Section~\ref{sec:resss}, we discuss the results based on the proposed metrics. We also perform qualitative analysis comparing the real and synthetic data distributions. Following a speaker independent cross-validation study we perform a cross-corpus with the two corpora having differences in speakers, recording conditions and annotators. Our aim was to observe the transferability of synthetic samples generated from a model trained on an external corporal to another corpora of interest. Finally we summarize our findings and mention some future avenues that could be worth exploring in Section~\ref{sec:conclusion}

\section{Related work}
\label{sec:related}
Speech emotion recognition is a widely researched topic and researchers in the past have leveraged the ability of generative models to learn a rich informative representations to build discriminative classifiers such as Gaussian mixture models (GMMs) and hidden Markov models (HMMs) \cite{el2011survey}. Chandrakala and Sekhar \cite{chandrakala2009combination} used Gaussian mixture models (GMMs) to model the distribution of feature sets extracted from utterances. They tried two different set-ups (i) each training sample is represented as a time-series of 34 dimensional feature vectors which are used to model a GMM. This results in M different GMMs for M training samples. Once the GMMs are trained, an M-dimensional score vector is computed for each utterance where each entry is the log-likelihood score obtained when the utterance is applied to one of the M GMMs. The score vectors are then used to train a support vector machine (SVM) which is evaluated on test samples. This approach is a simple way to obtain a fixed length representation for variable length utterances. Moreover, one can also see that similar utterances would generate similar log-likelihood scores resulting in similar score vectors which are then fed to SVM. (ii) Given that the previous model fails to capture the temporal dynamics of the utterance the authors tried a segment based approach. Each utterance was divided into a fixed number of segments. The set of feature vectors belonging to each segment were then modeled using a multivariate Gaussian distribution. The segment-wise feature vector is obtained by concatenating the entries in mean vector and the covariance matrix of the multivariate Gaussian. The final feature vector for an utterance used to train and evaluate the SVM classifier was obtained by concatenating the segment-wise feature vectors. The authors showed that segment based approach outperformed the score-vector based approach validating the importance of modeling temporal dynamics. Amer et al. \cite{amer2014emotion} proposed using hybrid networks by combining generative models that would draw rich representations of short term temporal dynamics and discriminative models which were used for classification of long range temporal dynamics using those representations. They experimented with using restricted boltzmann machines (RBMs) or conditional RBMs (CRBMs) as generative models while using an SVM or conditional random fields (CRFs) as discriminative models. They evaluated their models on three datasets and observed that the discriminative models trained with intermediate CRBM representations outperformed those trained using raw features. Latif et al. \cite{latif2017variational} leveraged the modeling capability of variational auto-encoders (VAEs) and conditional variational auto-encoders (CVAEs) to extract salient representations from log Mel filterbank coefficients for speech emotion recognition. VAEs are a class of auto-encoder based generative models that like GANs can generate synthetic data samples when provided with data points sampled from a simpler prior distribution $p_{\mathbf{z}}$. Along with minimizing the reconstruction error, the encoder output is made to resemble $p_{\mathbf{z}}$ by including a function in loss term that minimizes the Kullback- Leibler divergence between them. In case of CVAE, label information is provided while training and the latent representations are learned conditioned on both the input data and labels.

GAN based models have also been utilized to get meaningful representations of raw feature vectors for speech emotion recognition and related tasks. Since the discriminators in GANs are trained to discriminate between real and fake samples, one could use its intermediate layer representations obtained from raw feature vectors to train a classifier \cite{radford2015unsupervised}. Towards that end, Deng et al. \cite{deng2017speech} modelled a GAN whose generator was trained to output synthetic feature vectors that mimic the distribution of real acoustic feature vectors extracted from speech waveforms using the openSMILE toolkit \cite{eyben2010opensmile}. The intermediate layers of the discriminator were used to extract non-linear representations of the acoustic feature vectors. These were used to train an SVM to perform a 4-way classification of autism spectrum disorders. They showed an improvement in performance on their test set when they used the intermediate layer representations rather than the raw acoustic features. This indicates that the discriminators of a well trained GAN can extract meaningful representations from raw feature vectors. Chang and Scherer \cite{chang2017learning} followed a similar approach to obtain meaningful representations from spectrograms for valence level classification. They implemented a deep convolutional GAN architecture and used the activations from an intermediate layer of the discriminator for the final classification task. They reported better performance over a baseline model performing direct classification on spectrograms. Eskimez et al. \cite{eskimez2018unsupervised} used latent representations obtained from an auto-encoder based architectures and compared their performances for speech emotion recognition. Along with a denoising auto-encoder and a variational auto-encoder, they implemented an adversarial auto-encoder \cite{makhzani2015adversarial}, where the output of encoder is made to resemble a prior distribution using adversarial loss terms. They also implemented an adversarial variational bayes network which combined the advantages of variational auto-encoders and GANs. They reported that the representations obtained from an adversarial variational bayes network performed the best.

Besides these there have also been few other works utilizing GANs for speech emotion recognition that do not concern with extracting salient representations from raw feature vectors. Latif et al. \cite{latif2018adversarial} utilized GANs to build more robust speech emotion recognition systems by leveraging speech enhancement capability of GANs. They generated adversarial examples by adding noise (cafe, meeting or station noise) to actual training data  which could not be distinguished by human listeners from real examples in most cases. However, a classifier trained on real data couldn't classify the emotional class of the adversarial samples correctly. They trained a GAN to generate cleaner utterances from the corrupted utterances. They showed that the classifier was liable to make fewer miss-classifications when trained and evaluated on the cleaner data obtained from GAN than on perturbed data. Hence, inclusion of a GAN in their pipeline led to a more robust emotion recognition model. Han et al. \cite{han2018towards} proposed using a conditional GAN based affect recognition framework where the machine predicted labels were made to mimic the distribution of real labels. Their affect recognition framework consisted of a neural network classifier $NN1$ which when provided with the acoustic feature vector outputs the emotional class label. Another neural network $NN2$ was trained to distinguish between the ground truth labels and the output obtained from $NN1$. Hence, $NN1$ can be seen as a generator which generates 'fake' labels given the feature vectors and $NN2$ acts as a discriminator trying to differentiate between real and fake labels conditioned on the acoustic feature vector. The parameters of $NN1$ are updated based on a loss function which is a combination of the supervised cross-entropy loss term and a GAN based error term trying to confuse the discriminator $NN2$ between predicted and ground truth labels. Note that both these components would work towards making the predicted labels resemble ground truth labels. Like any GAN framework, the discriminator $NN2$ is updated so that it gets better at distinguishing between predicted and ground truth labels. They report that conditional adversarial training is helpful to improve speech emotion recognition showing GANs can be used to learn the label space distribution.

None of these works have however studied the generative capability of these models in greater detail. We have previously made attempts to investigate the ability of GAN based models to generate realistic feature vectors which can be used for speech emotion recognition \cite{sahu2018adversarial, sahu2018enhancing}. In this paper we explore this aspect in more detail by training three GAN based models to generate synthetic feature vectors mimicking the distribution of real acoustic feature vectors. We define the metrics to evaluate and compare the quality of the synthetic feature vectors generated using the three models. We provide visualizations comparing the distributions of real and synthetic data. Finally, we discuss the applicability of these synthetically generated feature vectors for speech recognition in low resource conditions. Note that since we are learning the distribution of feature vectors and not raw speech, the generated data could not be used for qualitative evaluation.

\section{Background on adversarial training}
\label{sec:background}
The purpose of a generative adversarial network (GAN) \cite{goodfellow2014generative} is to learn a complex distribution from a simpler distribution. Once trained the generator can be used to generate points from the complex distribution when fed with points from the simpler distribution. In this work, we attempt to train a generator that can produce synthetic high dimensional feature vectors (1582 dimensional vectors used for speech emotion recognition) from points belonging to simpler distributions. GANs consist of two modules : generator $G$ and a discriminator $D$ each of them having specific functions. Purpose of $G$ is to generate realistic data-points $G(\mathbf{z})$ when fed with samples $\mathbf{z}$ from a simpler distribution $p_{\mathbf{z}}$. Generally, $p_{\mathbf{z}}$ is chosen to be a Gaussian or a uniform distribution. Simultaneously, the discriminator is trained to be able to classify between the generated data-points $G(\mathbf{z})$ and real data-points $\mathbf{x}$. The final objective is to obtain a generator that can mimic real data distribution so that the discriminator is unable to differentiate between the generated data and real data. The loss function used to update the parameters of $D$ and $G$ is given by:
\begin{equation}\label{eq:gan_loss}
\begin{aligned}
\min_{G} \max_{D} V_\text{GAN}(D,G) = \mathbb{E}_{\mathbf{x} \sim p_{\text{data}}} [\log D(\mathbf{x})] + \\
\mathbb{E}_{\mathbf{z} \sim p_{\mathbf{z}}} [\log (1-D(G(\mathbf{z})))]
\end{aligned}
\end{equation}
Note that in the equation above $D(\mathbf{x})$ and $D(G(\mathbf{z}))$  denote the probabilities that $\mathbf{x}$ and $G(\mathbf{z})$ are recognized to be a real sample by the discriminator respectively. On the other hand parameters of $G$ should be updated such that it fools the discriminator into thinking that $G(\mathbf{z})$ comes from real data distribution. Hence, minimizing the loss term with respect to parameters of $G$ would push the value of $D(G(\mathbf{z}))$ closer to 1.
In practice, the parameters of $D$ and $G$ are updated in an iterative fashion with the parameters of one module kept frozen while the other one is updated. In each iteration, the number of updates to D and G could be different. 

In our experiments we use variations of adversarial auto-encoder network proposed by Makhzani et al. \cite{makhzani2015adversarial} for image classification and generation. We have explored their utility mainly for feature vector compression in \cite{sahu2018adversarial}. In adversarial auto-encoders we map the lower dimensional output of the bottleneck layer of an auto-encoder to a distribution $p_{\mathbf{z}}$. For a $N$-class classification problem, we can consider $p_{\mathbf{z}}$ to be a mixture of $N$ Gaussians, with each mixture component corresponding to a particular class. Once trained, points can be sampled from a particular mixture component and passed through the decoder to generate a synthetic data-point belonging to the corresponding class.

Another way to enforce clustering in the generated data is to use an infoGAN framework proposed by Chen et al. \cite{chen2016infogan} as opposed to specifying that $p_{\mathbf{z}}$ have $N$ components. In an infoGAN, the generator is fed with a conditional label vector $\mathbf{c} \sim p_{\mathbf{c}} $ along with $\mathbf{z} \sim p_{\mathbf{z}}$ to generate synthetic data-points $G(\mathbf{z},\mathbf{c})$. If we wish the generated data to have $N$ clusters, then $c$ can be a $N$ dimensional one-hot label vector. Along with the vanilla loss-term used for a GAN, an infoGAN additionally tries to maximize mutual information between generated sample distribution and the label distribution denoted by $I(\mathbf{c},G(\mathbf{z},\mathbf{c}))$ in the equation below.
\begin{equation}\label{eq:infogan_loss}
\min_{G} \max_{D} V_\text{infoGAN}(D,G) = V_\text{GAN}(D,G) - \lambda I(\mathbf{c},G(\mathbf{z},\mathbf{c}))
\end{equation}
Note that computing $I(\mathbf{c},G(\mathbf{z},\mathbf{c})$ requires us to compute the posterior $p(\mathbf{c}|G(\mathbf{z},\mathbf{c})$ which can be difficult. Chen et al. proposed a workaround by adding an auxiliary layer to their discriminator which classifies the generated synthetic vector. In other words it estimates the label $\mathbf{c'}$ given $G(\mathbf{z},\mathbf{c})$. The auxiliary layer is updated so that $\mathbf{c'}$ is as close to $\mathbf{c}$ as possible. Hence, they approximated the quantity $p(\mathbf{c}|G(\mathbf{z},\mathbf{c}))$ with the auxiliary layer output $q(\mathbf{c}|G(\mathbf{z},\mathbf{c}))$. More theoretical details can be found in \cite{chen2016infogan}. Denoting the approximation of $I(\mathbf{c},G(\mathbf{z},\mathbf{c}))$ with $Q(\mathbf{c},G(\mathbf{z},\mathbf{c}))$, the loss function to optimize now becomes
\begin{equation}\label{eq:infogan_app_loss}
\min_{G} \max_{D} V_\text{infoGAN}(D,G) = V_\text{GAN}(D,G) - \lambda Q(\mathbf{c},G(\mathbf{z},\mathbf{c}))
\end{equation}
This is the equation we used in our models utilizing the infoGAN framework. In our experiments, value of $\lambda$ was kept at 1. 
\subsection{InfoGAN vs vanilla GAN}
\begin{figure}[t]
\includegraphics[height=6cm, width=9cm]{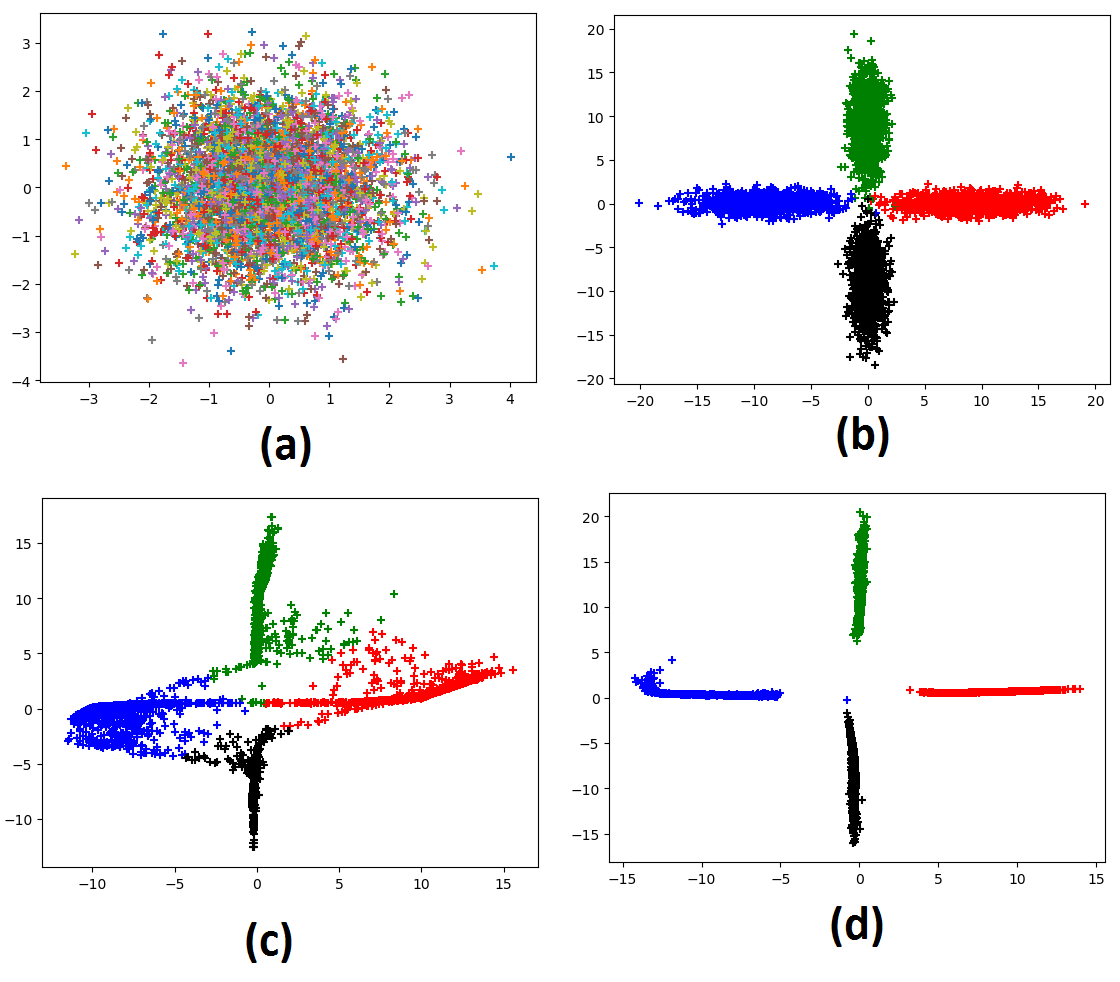}
\caption{Comparison of data generated using a trained vanilla GAN (c) and infoGAN (d). Source PDF is a 2D normal Gaussian distribution shown in (a) while target PDF is a mixture of 4 Gaussians as shown in (b). Note that the four clusters are quite separable in (d)}
\label{fig:info_vs_normal}
\end{figure}

To visualize how data generation varies between a vanilla GAN and infoGAN, we ran a simple experiment where we trained the two GAN models (for equal number of epochs) to learn a target Probability Distribution Function (PDF) from a source PDF. Our source PDF was a normal Gaussian distribution while target PDF was a mixture of 4 Gaussians with orthogonal means. Once trained, we sampled points from source PDF and fed it to the two GAN models. Figure~\ref{fig:info_vs_normal} shows the differences in the distribution of the generated data-points. We note that the generated data from a infoGAN has high inter-cluster variance with no overlap between samples belonging to different clusters. However, the inter-class variability is low as all the samples lie along a straight line. A vanilla GAN on the other hand doesn't exhibit these properties. Hence, the mutual information based loss function focusses more on inter-cluster separability than intra-cluster variance. This is something we should keep in mind as we will discuss more on this in later sections.

We now discuss our experimental set up explaining the architectures of the models used and the training methodology in more detail.
\section{Experimental setup}
\label{sec:exp}
In this section we explain the databases and the architectures and training methodology of the three GAN based models trained by us to generate synthetic samples. We only used data from four emotion classes namely angry, sad, neutral and happy to train the models. We used the openSMILE toolkit to extract 1582 dimensional `emobase' feature set from raw speech waveforms which was used in our experiments. This feature set consists of various functionals computed for spectral, prosody and energy based features. Similar features have been previously used for emotion classification \cite{el2011survey}. We then define the metrics used by us to compare the synthetic data generation capability of the three different models. We perform an in-domain cross validation analysis where we train the GAN using samples from training split. We used the metrics defined by us to compare the synthetic data distribution with the data distribution in training and validation splits. This is followed by a cross-corpus analysis where the synthetic data generated from a GAN trained on one corpus was compared with data belonging to a different corpus.

\subsection{Datasets}
We use IEMOCAP and MSP-IMPROV datasets for our analysis. These datasets are one of the larger datasets used by the emotion recognition community \cite{lotfian2017building}. Another desirable property of IEMOCAP which is used to train the GAN models is the more balanced distribution of emotion labels compared to other datasets.
\subsubsection{IEMOCAP}
Interactive  Emotional  Dyadic  Motion  Capture (IEMOCAP) dataset \cite{busso2008iemocap} consists of five sessions with dyadic affective interactions. In each session, two actors act out scenarios which are either scripted or improvised. No two sessions have the same actor participating in them. This enabled us to perform a five fold leave-one-session out cross-validation analysis on IEMOCAP. The conversations have been  segmented into utterances  which  are then  labeled  by three  annotators  for emotions such as happy, sad, angry, excitement and, neutral. For our experiments, we only use utterances for which we could obtain a majority vote and assign that as the ground truth label. We used approximately 7 hours of data from the dataset which amounts to  5530  utterances : neutral  (1708),  angry  (1103),  sad  (1083),  and  happy  (1636).
\subsubsection{MSP-IMPROV}
MSP-IMPROV \cite{busso2017msp} has actors participating in dyadic conversations across six sessions and like IEMOCAP they also have been segmented into utterances. But unlike IEMOCAP, it also includes a set of pre-defined 20 target sentences that are spoken with different emotions depending on the context of conversation. There  are  7798  utterances belonging to the same four emotion classes. The class distribution is unbalanced with the number of utterances belonging to happy/neutral class (2644 and 3477 respectively) more than three times that of angry/sad (792 and 885 respectively). We used MSP-IMPROV to perform a cross-corpus study using it as a test set while IEMOCAP was used as training set.
\subsection{GAN architectures employed}
\label{sec:archi}
We used auto-encoder and GAN based models where the bottleneck layer of the auto-encoder learns lower dimensional code vectors from higher dimensional feature vectors. The lower dimensional encoding space is made to resemble a simple prior $p_{\mathbf{z}}$ by using a GAN based training framework. Points from this lower dimensional subspace are then sampled and fed to the decoder of the auto-encoder to get synthetic feature vectors. Size of the the bottleneck layer in our architectures was decided so that we could obtain lower dimensional encodings without sacrificing much of the discriminability present in the actual higher dimensional feature vectors. To quantify this, we ran a cross-validation experiment on IEMOCAP for a 4-way emotion classification by training an SVM using raw features as well as the lower dimensional encodings obtained by feeding the trained models with raw features similar to the set up described in \cite{sahu2018adversarial}. The drop in accuracy would give us an idea of the amount of discriminability retained by the lower dimensional encodings. Using the architectures mentioned in this paper, we noticed that the classification accuracy dropped by only a couple of percentage points when lower dimensional code-vectors were used instead of raw feature vectors suggesting they indeed retain most of the discriminability. We employ three models $M1$, $M2$ and $M3$ for the task of synthetic data generation from a simpler distribution $p_{\mathbf{z}}$. All of them consist of fully connected layers. We describe them in more detail below.
\begin{figure*}[t]
\includegraphics[height=5cm, width=18cm]{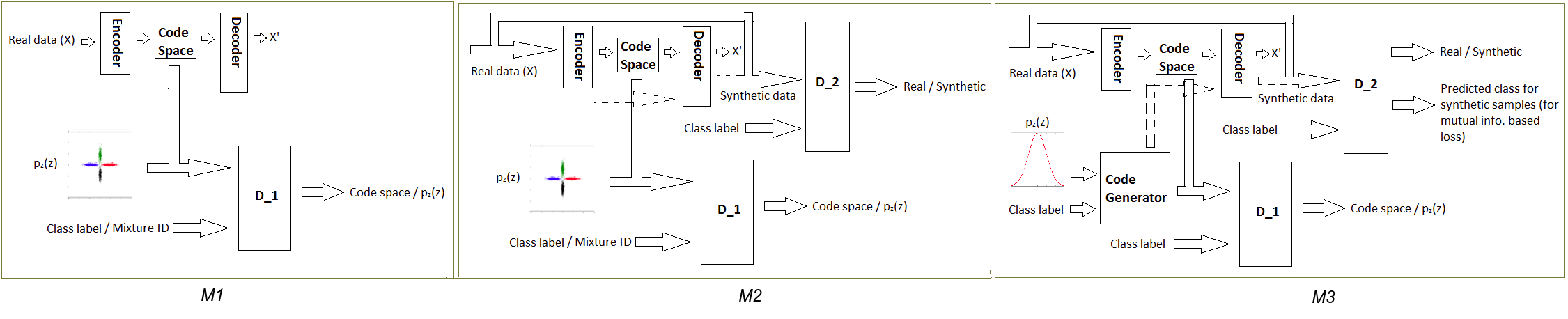}
\caption{Architectures for $M1$ (left), $M2$ (center) and $M3$ (right). Note that there are two discriminators in $M2$ and $M3$, one to learn the encoding space and one to generate data samples. While in $M1$ and $M2$, the encoding space is pre-defined to be a mixture of 4 maximally separated Gaussians, in case of $M3$ it is being learned from the training data provided using a code generator block}
\label{fig:AAEdgan}
\end{figure*}

\begin{table*}[t]
\caption{Architecture of various components of models $M1$, $M2$ and $M3$. Except the bottleneck layer and output layer in auto-encoders (linear activations), output layer of discriminators $D_1$ and $D_2$ (sigmoid activations) and output layer of code-generator in $M3$ (linear activation) all the other layers had ReLU activations.}
\begin{tabular}{l|l|l} \hline
\textbf{Component}  & \textbf{Model} & \textbf{Architecture}  \\ \hline
Auto-encoder & $M1$ and $M2$ & $1582\rightarrow{}1000\rightarrow{}500\rightarrow{}100\rightarrow{}\textbf{2}\rightarrow{}100\rightarrow{}500\rightarrow{}1000\rightarrow{}1582$ \\
(Enc/Dec with bottle-neck layer in bold) &$M3$&$1582\rightarrow{}1000\rightarrow{}700\rightarrow{}300\rightarrow{}\textbf{256}\rightarrow{}300\rightarrow{}700\rightarrow{}1000\rightarrow{}1582$\\\hline
Discriminator $D\_1$& $M1$ and $M2$ & $6\rightarrow{}1000\rightarrow{}500\rightarrow{}100\rightarrow{}1$\\
&$M3$ & $260\rightarrow{}1000\rightarrow{}500\rightarrow{}100\rightarrow{}1$\\\hline
$p_{\mathbf{z}}$& $M1$ and $M2$ & Mixture of four 2D Gaussians with orthogonal means\\
&$M3$&20 dimensional normal distribution \\\hline
Code generator (CG) & $M3$ & $24\rightarrow{}140\rightarrow{}256$\\\hline
Discriminator $D\_2$& $M2$ and $M3$ & $1586\rightarrow{}1000\rightarrow{}500\rightarrow{}100\rightarrow{}1$\\\hline
Auxiliary layer (AUX)&$M3$&$[1586\rightarrow{}1000\rightarrow{}500\rightarrow{}100]\rightarrow{}128\rightarrow{}4$\\
(Layers within brackets are shared with $D\_2$)&&\\\hline
\end{tabular}
\label{tab:archi_m}
\end{table*}
\begin{itemize}
\item $M1$: $M1$ is an adversarial auto-encoder with $p_{\mathbf{z}}$ as mixture of four 2D Gaussians with orthogonal means and same mixture weights. This also meant the bottleneck layer can have two neurons. We trained the model to match the distribution of code vectors (output of bottleneck layer) to that of $p_{\mathbf{z}}$. While training, we used the label information in the form of one-hot vectors to match each emotion category to a particular mixture component. We match the distributions using a GAN framework with a discriminator $D\_1$ trying to distinguish between code-vectors and samples obtained from $p_{\mathbf{z}}$. In this framework, the encoder can be viewed as a generator which generates lower dimensional encodings when provided with real feature vectors. Once trained the encodings would match the distribution $p_{\mathbf{z}}$ with each emotion category being mapped to one mixture component. We can also sample points from $p_{\mathbf{z}}$ to feed the decoder thereby generating synthetic feature vectors. 
\item $M2$: Note that the decoder in $M1$ does not receive any feedback to update its parameters from a GAN framework trying to match its output to real feature vectors. In model $M2$, we included a second discriminator $D\_2$ to the architecture $M1$ that can distinguish between decoder samples and real feature vectors. Decoder parameters can now be updated taking advantage of the information provided by $D\_2$. Label information was used for matching the distribution of real feature vectors belonging to a specific emotion class to the distribution of synthetic vectors generated from a specific component of $p_{\mathbf{z}}$ . In this case, decoder acts as the generator trying to generate synthetic feature vectors when provided with samples from $p_{\mathbf{z}}$. 
\item $M3$: Note that in both the architectures described above, clustering of synthetic data into four emotion classes is ensured by specifying a $p_{\mathbf{z}}$ having four components with orthogonal means. In model $M3$ we enforce this clustering by implementing an infoGAN framework. At the same time clustering of code vectors into four classes was data-driven. \cite{wang2018learning} has explored similar models for synthetic image generation. Since the code-space clustering is data-driven we had to increase the dimension of bottleneck layer to 256 neurons to retain the discriminability present in raw features as described before. $p_{\mathbf{z}}$ is modeled to be 20 dimensional normal distribution. When fed to a code generator network the output spanned a 256 dimensional space which was matched to the code-vector distribution. Once trained, points are sampled from $p_{\mathbf{z}}$ and provided as input to the code generator followed by the decoder network generating synthetic feature vectors. The architecture is shown in Figure~\ref{fig:AAEdgan}. Note that since we are using an infoGAN framework to generate synthetic data, conditional information $\mathbf{c}$ is being provided in the form of one-hot label vectors to the code generator along with points from $p_{\mathbf{z}}$. We take $\mathbf{c}$ to be sampled from a discrete uniform distribution implying the synthetic feature vectors are equally likely to belong to any of the four classes. Discriminator $D\_2$ had an auxiliary layer predicting the class of the synthetically generated samples. This output was used to approximate the conditional distribution of the labels given the synthetic feature vectors. 
\end{itemize}
The three architectures are shown in Figure~\ref{fig:AAEdgan} and more details for various components of the models are provided in Table~\ref{tab:archi_m}.

\subsection{Training methodology}
\label{sec:trainingm}
In this section we outline the methodology used to train our models. As mentioned before GANs have a discriminator and a generator playing a min-max game trying to fool each other. At the end of training, the loss curves obtained from discriminator and generator networks should converge implying that the GAN has achieved an equilibrium with the generator producing realistic enough samples to confuse the discriminator. 
Usually a generator's job to learn a complex distribution is more difficult than a discriminator's job to simply classify between real and generated data. Hence updating a generator more number of times than a discriminator in each iteration can be helpful. However, a weak discriminator that doesn't do a good job of classifying real and generated data is undesired as the discriminator's output drives the the generator to produce more realistic samples. Hence, a careful tuning regarding the number of updates to each module is required.

Below we mention the training steps for our GAN based models. We used stochastic gradient descent as the optimizer to update the model parameters. The learning rate and other training parameters for the various components are tuned so that the discriminator and the generator errors converge as the training progresses. Note that while for $M1$ we have one GAN framework to match the coding space to $p_{\mathbf{z}}$, for $M2$ and $M3$ we have an additional GAN framework to match the distribution of synthetic samples coming out of the decoder to that of real feature vectors. The former task is easier because the feature vectors lie in a higher dimensional space compared to the code-vectors. For all three models we start off with updating the auto-encoder's weights to minimize reconstruction error followed by training the GAN framework to match the code-vector distribution with that of samples obtained from $p_{\mathbf{z}}$ . Since $M2$ and $M3$ have an extra component in the form of discriminator $D\_2$, additional steps were implemented to train the GAN framework trying to match decoder's output with real feature vectors. For a particular batch of the training sample, the different components in $M1$, $M2$ and $M3$ were updated in the following steps:
\begin{itemize}
\item Step 1 - Update auto-encoder wights: Weights of the auto-encoder (encoder and decoder) are updated based on a reconstruction loss function. We considered mean squared error to be the reconstruction loss function. Hence for a real data sample $\mathbf{x}$ and its reconstruction $\mathbf{x'}$, the auto-encoder weights were updated to minimize $\|\mathbf{x}-\mathbf{x'}\|_2^2$. We used a learning rate (LR) of 0.001 with a momentum of 0.9 for $M1$ and $M2$. LR was kept the same for $M3$ but no momentum was used.

\item Step 2 - Update $D\_1$ weights: Real data-points are transformed by the encoder. An equal number of points are sampled from $p_{\mathbf{z}}$. In case of $M3$, the sampled points are also passed through the code generator (CG) along with the conditional labels $\mathbf{c}$. Weights of the discriminator ($D\_1$ in pictures) are updated to minimize cross-entropy to distinguish between encoded samples and samples obtained/derived from $p_{\mathbf{z}}$. Note that label information is also provided to discriminator. Let us consider a real sample $\mathbf{x}$ and $\mathbf{c_x}$ to be the one-hot vector denoting its class. Let $\mathbf{z}$ be a sample obtained from $p_{\mathbf{z}}$ and $\mathbf{c_z}$ be the one hot vector denoting the mixture component it was sampled from in case of $M1$ and $M2$.  Assuming the ground truth label when $D\_1$ gets the code-vectors $enc(\mathbf{x})$ as input is 1 and it's 0 when $D\_1$ is provided with samples obtained/derived from $p_{\mathbf{z}}$, loss function minimized to update the discriminator's parameters is given by:
\begin{equation}\label{eq:gan_loss2}
\text{\bf M1, M2:}\; \log(D(\text{enc}(\mathbf{x}),\mathbf{c_x}))-\log(1-D(\mathbf{z},\mathbf{c_z}))
\end{equation}
\begin{equation}\label{eq:gan_loss3}
\text{\bf M3:}\; \log(D(\text{enc}(\mathbf{x}),\mathbf{c_x}))-\log(1-D(\text{CG}(\mathbf{z},\mathbf{c}),\mathbf{c}))
\end{equation}
LR of 0.1 was used for $M1$ and $M2$ while it was 0.01 for $M3$.

\item Step 3 - Update encoder weights: We then freeze the discriminator (D\_1) weights. The weights of encoder are updated based on its ability to fool the discriminator. Hence, for a real sample $\mathbf{x}$, the ground truth label when $D\_1$ gets the code-vectors $enc(\mathbf{x})$ as input is now 0. Loss function minimized to update the encoder's parameters is given by $-\log(1-D(enc(\mathbf{x}),\mathbf{c_x}))$. LR of 0.1 was used for $M1$ and $M2$ while it was 0.01 for $M3$.
\end{itemize}
For $M2$ and $M3$ there were two additional steps as mentioned below to match the decoder's output with the distribution of real feature vectors.
\begin{itemize}
\item Step 4 (Only for M2/M3): Points $\mathbf{z}$ are sampled from $p_{\mathbf{z}}$ and fed to decoder in case of $M2$ or to code generator + decoder along with a class label in case $M3$. Weights of the discriminator ($D\_2$ in pictures) are updated to minimize cross-entropy to classify between synthetic samples and real samples.  Assuming the ground truth label when $D\_2$ is fed with the decoder's outputs as input is 1 and it's 0 when $D\_2$ is provided with real samples $\mathbf{x}$, loss function minimized to update the discriminator's parameters is
\begin{equation}\label{eq:gan_loss4}
\text{\bf M2:}\; \log(D(dec(\mathbf{z}),\mathbf{c_z}))-\log(1-D(\mathbf{x},\mathbf{c_x}))
\end{equation}
\begin{equation}\label{eq:gan_loss5}
\text{\bf M3:} \;\log(D(dec(CG(\mathbf{z},\mathbf{c})),\mathbf{c}))-\log(1-D(\mathbf{x},\mathbf{c_x}))
\end{equation}
LR of 0.0001 was used for both $M2$ and $M3$.
\item Step 5 (Only for M2/M3): We then freeze the discriminator ($D\_2$) weights. The weights of decoder (in case of $M2$) or code generator + decoder (in case of $M3$) are updated based on its ability to fool the discriminator. In case of $M3$, an additional loss term based on mutual information $Q$ is also considered to update the parameters.
\begin{equation}\label{eq:gan_loss6}
\text{\bf M2:}\log(1-D(\text{dec}(\mathbf{z}),\mathbf{c_z}))
\end{equation}
\begin{equation}\label{eq:gan_loss7}
\begin{aligned}
\text{\bf M3:}\log(1-D(\text{dec}(CG(\mathbf{z},\mathbf{c})),\mathbf{c})) \\
-Q(\mathbf{c},\text{dec}(CG(\mathbf{z},\mathbf{c})))
\end{aligned}
\end{equation}
Furthermore, the auxiliary layer (AUX) weights are also updated busing the mutual information based loss term. LR of 0.001 was used for both $M2$ and $M3$.
\end{itemize}
Note that the learning rate used to update $D\_2$ parameters is less than that used to update the decoder/code generator + decoder so as to balance out the learning abilities of the generator and discriminator in this GAN framework. Also, the generators were trained for two epochs for every single epoch of training $D\_2$. 
\subsection{Evaluation metrics}
Once the models are trained till the discriminator and generator errors converge, we can sample points from $p_{\mathbf{z}}$ and feed it to the decoder (in case of $M1$ and $M2$) or code generator followed by decoder (in case of $M3$) to generate synthetic feature vectors. For $M1$ and $M2$, the corresponding label of the generated feature vectors is the same as the mixture id of the Gaussian component from which $\mathbf{z}$ was sampled. For $M3$, the synthetic feature vectors are assigned to the class denoted by the label vector $\mathbf{c}$ being fed to code generator with respect to which we maximize the mutual information of the generated feature vectors. Hence, using our trained GAN models, we are able to sample points from the distribution $p(\mathbf{x_{synth}},y_{synth})$ where $\mathbf{x_{synth}}$ represents the synthetic feature vectors and $y_{synth}$ represents their labels. To evaluate the effectiveness of our models, we need to compare the distribution $p(\mathbf{x_{synth}},y_{synth})$ with real distribution $p(\mathbf{x_{real}},y_{real})$. So far, a standardized set of metrics that can quantify the similarity between real and fake samples is not available. To address this, we suggest three metrics and evaluate the models on them. We define these evaluation metrics below.
\subsubsection{Metric 1: Testing accuracy on synthetic data with a classifier trained on real data}\label{sec:synth_test}
The objective of this experiment is to assess the similarity between real and synthetic data by using a model trained on real data to classify synthetic data. This would give us an idea about the quality of the synthetic data. A higher accuracy would suggest the generated distribution $p(\mathbf{x_{synth}},y_{synth})$ is very much similar to the real distribution $p(\mathbf{x_{real}},y_{real})$. However, it may so happen that the variance within samples belonging to the same class is low i.e. they do not capture the full distribution of the modeled class. On the other hand a lower accuracy would suggest the real and synthetic data samples comes from relatively different distributions. It doesn't necessarily imply that synthetic data is bad quality. It may so happen that the synthetic data is generating meaningful samples not represented in the real dataset. 
\subsubsection{Metric 2: Testing accuracy on real data with a classifier trained on synthetic data}\label{sec:synth_train}
In this experiment we evaluate the performance of a model trained on synthetic data to classify the test set consisting of real data. A high accuracy indicates that the generative models produce samples that are good representations of all the classes. This measure would reflect the diversity of synthetic data. For example, the classifier trained using synthetic samples form a GAN model that's liable to mode collapse \cite{arjovsky2017wasserstein} would perform poorly because the training set will have samples from only a few classes. Also we can verify if a GAN based model generates meaningful samples because then it can be used to train a classifier to classify real data even if they are not explicitly present in the real dataset. 
\subsubsection{Metric 3: Using Fretchet Inception Distance (FID) metric}\label{sec:FID}
This metric derives its name from the Inception network \cite{szegedy2015going}. It is a deep convolutional neural network model trained on millions of images for the purpose of image classification. Inception net had reported state of the art for classification and detection in the ImageNet Large-Scale Visual Recognition Challenge 2014. Since, its been trained on a huge dataset it is assumed that the network is generalizable enough and the intermediate layers produce meaningful activations helpful for image classification. Researchers have used this for evaluating the quality of synthetic samples generated by GANs \cite{heusel2017gans}. To compute FID, inception network is used to get intermediate layer activations for real and synthetic dataset. Then we compute the statistics : mean $\mu$  and covariance $\Sigma$ for those activations over all the samples present in the corresponding fake and synthetic datasets. The FID between the real images x and generated images g is computed as:
\begin{equation}\label{eq:fid}
FID(x,g) = \|\mu_x-\mu_g\|_2^2 + Trace(\Sigma_x + \Sigma_g - 2(\Sigma_x\Sigma_g)^{0.5})
\end{equation}
Hence, if the generated and real images come from similar distributions, FID will be lower. Also note that while in previous two metrics we are comparing the joint distributions $p(\mathbf{x_{synth}},y_{synth})$ and $p(\mathbf{x_{real}},y_{real})$, FID compares the marginals $p(\mathbf{x_{synth}})$ and $p(\mathbf{x_{real}})$.

Unfortunately, there isn't a deep network like Inception net for speech emotion recognition which has been trained on a whole lot of data to be genralizable enough to extract meaningful activations when provided with raw feature vectors. Hence, we used a neural network with fully connected layers trained on IEMOCAP and derive the activations from its intermediate layer for FID computation. Note that GAN models generating data similar to IEMOCAP will have a lower FID thereby judging them to be "better" models by this metric. However, as mentioned before a lower score doesn't necessarily mean worse models because synthetic dataset can have meaningful samples not represented in the limited IEMOCAP dataset.

\section{Results}
\label{sec:resss}
In this section we evaluate the GAN based models based on the characteristics of the generated synthetic data. We first execute an in-domain 5-fold cross-validation analysis on IEMOCAP. As has been mentioned before each of the 5 sessions in the dataset have different speakers so by doing a leave-one-session out cross-validation we ensure a speaker independent analysis. We also perform a visual analysis to compare the distributions of real and synthetic datasets for one of the cross-validation splits. Finally, we do a cross-corpus study where we train a GAN model using IEMOCAP and compare the synthetic data generated to real feature vectors obtained from utterances in MSP-IMPROV. Along with computing the above three metrics, we simulate a low resource condition to find out if using synthetic data along with limited real data to train a model aids in emotion classification. The purpose of this experiment was to judge the transferability of knowledge provided by synthetic vectors across two corpora. If that was the case appending the real dataset from a source (IEMOCAP) distribution with synthetic data obtained from GANs trained on the source distribution would aid the classification of test samples belonging to target (MSP-IMPROV) distribution. We have previously seen in \cite{sahu2018adversarial} and \cite{sahu2018enhancing} how appending real dataset with synthetic dataset improves the in-domain classification only by a few (1\% approx) percentages. Here we undertake a more in-depth analysis of how cross-corpus emotion recognition is affected.
\subsection{In-domain cross-validation experiments}
\begin{figure*}[t]
\includegraphics[height=13cm, width=17cm]{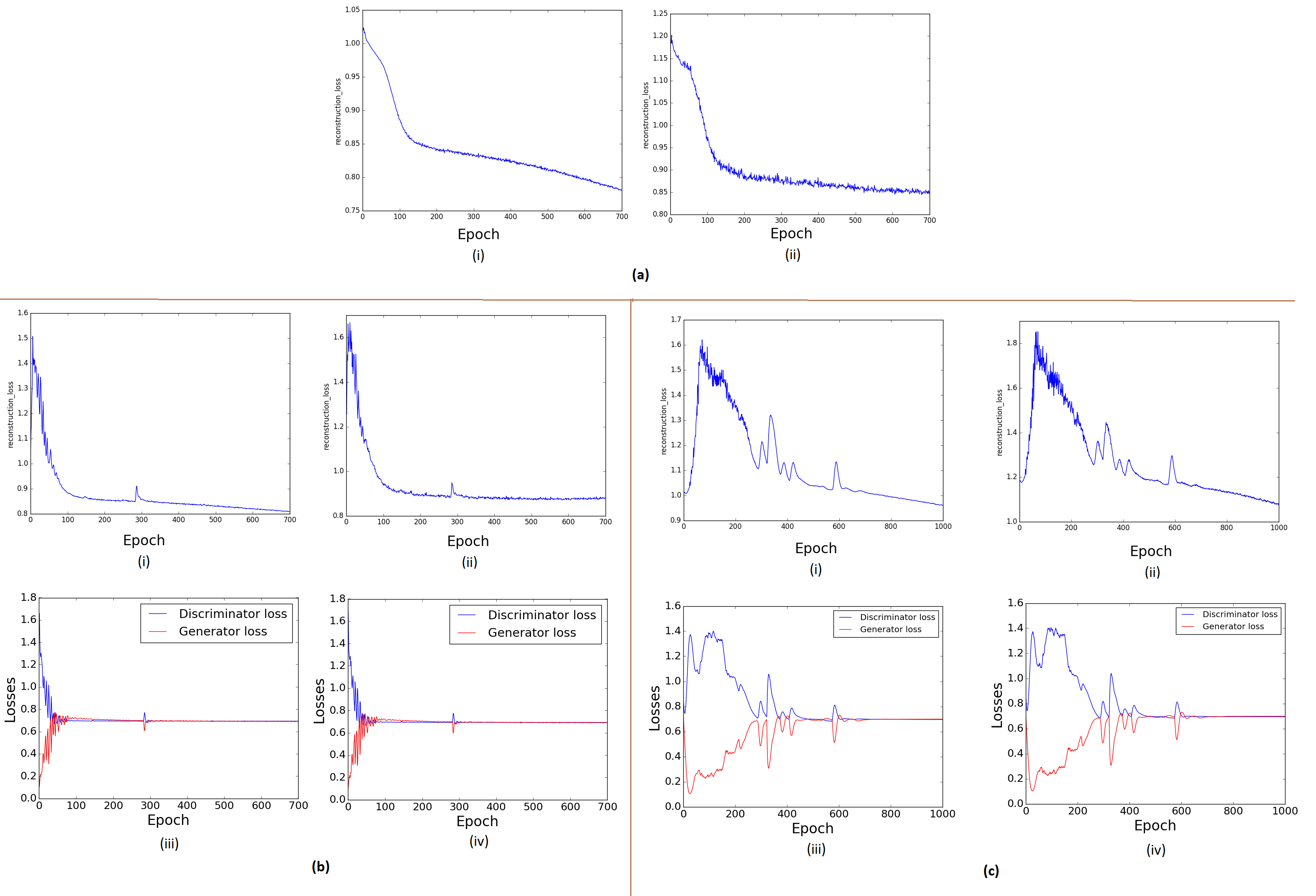}
\caption{Reconstruction or adversarial errors (discriminator’s (blue) and generator’s (red) errors) for one of the cross-validation splits (a) $M1$ (b) $M2$ (c) $M3$. a(i), b(i,iii), c(i,iii) belong to training set while a(ii), b(ii,iv), c(ii,iv) belong to validation set. Note how the discriminator and generator's errors are converging indicating GANs have reached a equilibrium state. Also the trends are similar for training and validation splits indicating the models generalize well.}
\label{fig:errors}
\end{figure*}

Figure~\ref{fig:errors} shows the reconstruction loss curves for for models $M1$, $M2$ and $M3$ trained on four sessions for a given cross-validation iteration. We also show how the discriminator ($D\_2$) and generator's losses of the data-generating part of the models change as the training progresses. Note that the errors converge for both training and validation splits indicating that the GAN errors converge for both of them. Once trained, we sample points from $p_{\mathbf{z}}$, pass it through the respective decoders and get the points $(\mathbf{x_{synth}},y_{synth})$. In case of $M1$ and $M2$, we are equally likely to sample points from any of the four modes of the mixture PDF $p_{\mathbf{z}}$. For $M3$, the conditional label vectors $\mathbf{c}$ are sampled from a uniform distribution. Since there is almost equal representation from all classes, the synthetic dataset will be balanced. For each CV split, the generated data can either be compared to the training set (set-1) or the validation set (set-2).  Post the GAN model training, we compute metrics 1 and 2 with the real datapoints sourced either from training set (set-1) or validation set (set-2).  We train separate SVM classifier on the real and synthetic dataset to compute metrics 1 and 2, respectively. The model hyperparameters are tuned on the respective training set. We report the average unweighted accuracy (UWA) over the five cross-validation splits. We generate 6000 synthetic data-points, approximately the same number of data-points as the IEMOCAP set used in experiments. We show the results for the different train and test conditions in Table~\ref{tab:synth_res} (chance accuracy = $\frac{1}{4} = 25\%$).
\begin{table}[t]
\centering
\caption {Cross-validation accuracies (\%) obtained using different combinations of data-sets for training and evaluating an SVM classifier. Datasets used to train and test the SVM classifier are denoted as Tr. and Te. in the table respectively. Set-1 refers to the training set of a cross-validation split used to train the GAN model while set-2 refers to the validation set.}
\begin{tabular}{@{}l|c|c|c|c@{}} 
 Models&Metric 1  & Metric 2  & Metric 1  & Metric 2 \\
 &Tr. : Set-1 & Te. : Set-1  & Tr. : Set-2  & Te. : Set-2 \\ \hline
$M1$      &85.60& 48.58 & 72.52 &45.89  \\
\hline
$M2$      &\textbf{88.41}& 49.91 & \textbf{74.75} &46.96  \\
\hline
$M3$     &55.20& \textbf{52.24} &  45.63&\textbf{51.58}  \\
\hline
\end{tabular}
\label{tab:synth_res}
\end{table}

It can be seen that results obtained using set-1 to train/test the SVM classifier are better than if set-2 were used instead. This is expected as set-1 was used to train the GAN models and hence the generated data is expected to be similar to set-1. Note that the set-2 contains a different set of speakers, adding to the mismatch. It can be observed that the accuracies obtained for $M2$ are better than that obtained for $M1$. This indicates that $M2$ generated synthetic samples have more similarity to samples obtained from real data distribution than $M1$. Note that the decoders are trained differently for the two models. While in case of $M1$ decoder parameters are updated based only on reconstruction loss, in case of $M2$ the parameters are updated based on an additional adversarial loss that determines how close it is to real data. We hypothesize that this extra update is what leads to better synthetic sample generation by $M2$ that also generalizes better for unseen speakers. Another interesting thing to note is the characteristic of synthetic data generated using $M3$. When used as test set, the classifiers trained on real data are unable to classify them as good as they classify samples generated from $M1$ and $M2$. This indicates unlike $M1$ and $M2$, $M3$ produces synthetic data samples that are not represented in the real data distribution and hence a classifier trained on real data fails to recognize their classes accurately. However, training a classifier on synthetic data obtained from $M3$ performs better at classifying real data-points than a classifier trained on samples generated from $M1$ and $M2$. This indicates that data generated using the model $M3$ has more diverse samples than data generated using $M1$ and $M2$. These differences arise due to the difference in training procedures and the prior $p_{\mathbf{z}}$ from which points are sampled to be fed into the decoder to generate synthetic samples. While in case of $M1$ and $M2$ it is pre-defined to be a mixture of four Gaussians, in case of $M3$ the GAN model learns it during training by maximizing the mutual information between generated samples and the conditional label vector used to generate them. Also, the coding space of $M3$ has more dimensions (256) than $M1$ and $M2$ (2) which could provide the decoder with a wider range of input samples which probably leads to diverse synthetic data-points. It is interesting to note that even though data generated using $M3$ doesn't resemble the samples in the real dataset used to train it, they still contain enough meaningful samples which can help recognize the classes of real data points. Next, we evaluate the generated data from the three different models using the FID metric.

FID metric uses the intermediate layer's outputs for obtained from a trained neural network for real and synthetic data sets. For our purposes we used a fully connected neural network with 4 hidden layers of 64 neurons each and regularized linear (ReLU) activation followed by an output layer of 4 neurons (each neuron corresponding to an emotion class) with softmax activation. The input layer had 1582 neurons corresponding to the dimension of the feature vectors used to train it for emotion recognition. The network was trained for 30 epochs on the entire IEMOCAP dataset containing samples from the four emotional classes of interest. Once trained, the weights were frozen and output of the third hidden layer was obtained for real and synthetic samples. Then the statistics $\mu$  and $\Sigma$ were computed for real and synthetic datasets and these were used to calculate the FID metric according to equation ~\ref{eq:fid}. The metric averaged over the 5 cross-validation splits is shown in Table~\ref{tab:fid_res}. As can be seen the scores are lowest for $M2$ closely followed by $M1$ indicating the synthetic data generated by these models are more similar to the real IEMOCAP data than those generated by $M3$. This confirms our findings presented in Table~\ref{tab:synth_res} where we observed that an SVM classifier trained on real data does a better job of classifying synthetic samples generated from $M1$ and $M2$. This doesn't necessarily mean that $M3$ is worse than the other two models as we saw from Table~\ref{tab:synth_res} that it generates more diverse and meaningful samples that might not be represented in the limited IEMOCAP dataset. Next, we show some visualizations to qualitatively analyze the quality of synthetic data generated from the three models. 
\begin{table}[t]
\centering
\caption {FID metric when synthetic data from the three models was compared to real data distribution. Note that lower FID means that the distributions are more similar.}
\begin{tabular}{@{}l|c|c|c@{}} \hline
 Models&$M1$  & $M2$ & $M3$ \\ \hline
FID      &14.52& \textbf{13.24} & 33.99  \\
\hline
\end{tabular}
\label{tab:fid_res}
\end{table}
\subsubsection{t-SNE analysis of generated data}\label{sec:tsne}
\begin{figure}[t]
\includegraphics[height=9cm, width=9cm]{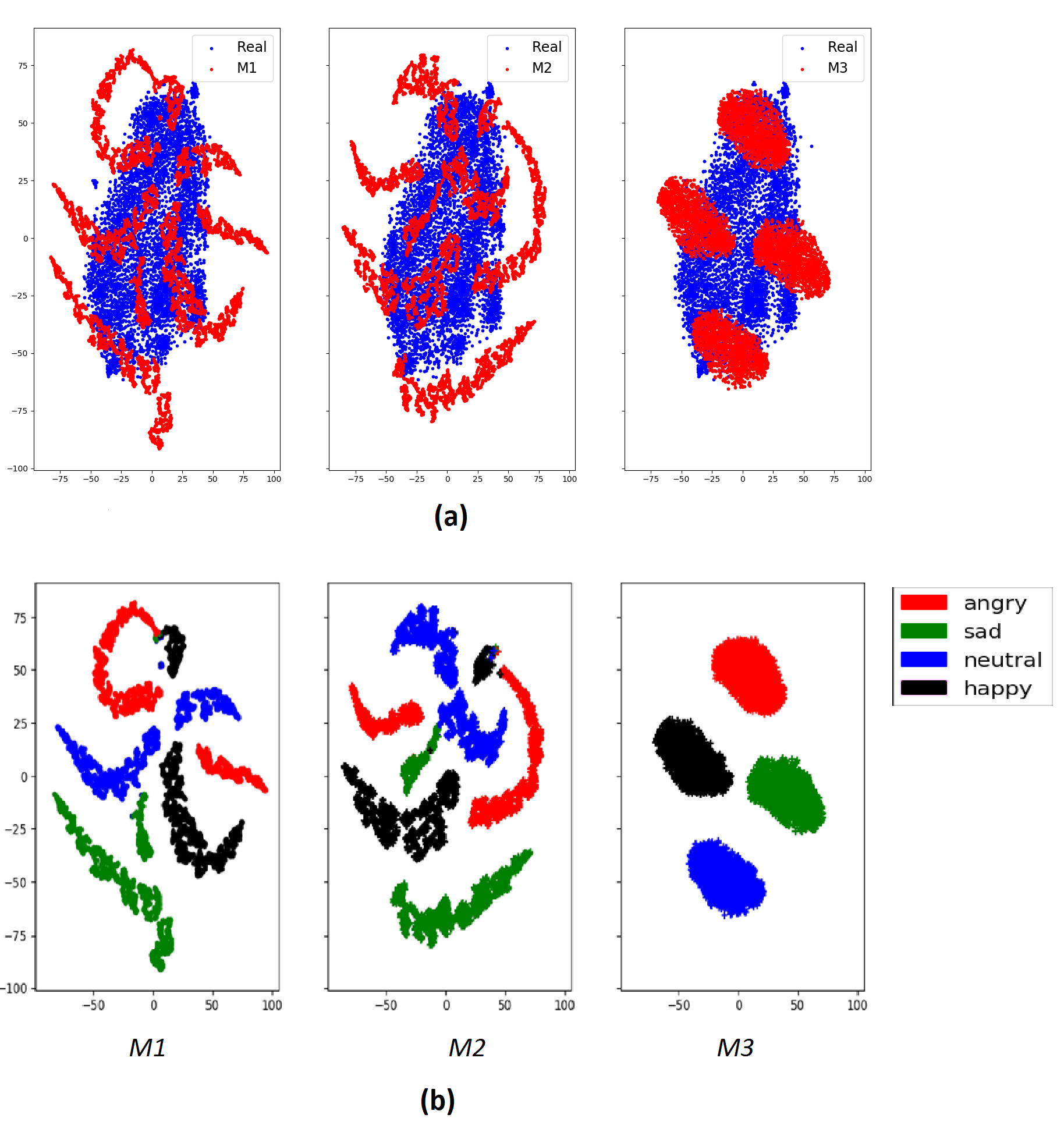}
\caption{(a) Comparison of t-SNE embeddings of synthetic feature vectors generated using $M1$, $M2$ and $M3$ with the embeddings of real IEMOCAP data (b) Class-wise clustering of synthetic data generated using the three models.}
\label{fig:tsne_synth}
\end{figure}

In Figure~\ref{fig:tsne_synth}(a), we compare the 2D t-distributed stochastic neighborhood embeddings (t-SNE) plots of 1582-D synthetic feature vectors generated using models $M1$, $M2$ and $M3$ with each other and with that of real data. We see that for all three models, the majority of the synthetic data embeddings lie in the space defined by the real data which indicates that the GAN models are indeed capturing the underlying feature vector distribution to some extent. Additionally it can be seem that the distributions of synthetic data generated from models $M1$ and $M2$ resemble each other. This is because the prior $p_{\mathbf{z}}$ used to generate the synthetic data is same in both these models and different from that used in $M3$. In fact data generated using $M3$ form four separate clusters, each corresponding to an emotion Figure~\ref{fig:tsne_synth}(b). This again points to the observation made in Section~\ref{sec:background} that InfoGAN based models try to increase the inter-class variability while not giving as much attention to intra-class variability. This can explain the results in Table~\ref{tab:synth_res} where we saw SVM trained on synthetic data generated using $M3$ giving us better accuracies in classifying real data than those trained on data generated from $M1$ and $M2$. The greater inter-class variability in synthetic training data leads to formation of more separable hyper-planes when training an SVM. On the other hand, since $M1$ and $M2$ don't focus on maximizing inter-class variability, less restrictions are imposed on them when they try to capture the real underlying data distribution. This leads to GAN models that generate synthetic data samples closer to the real data samples used to train the corresponding models. Such synthetic samples can be classified with greater accuracy using a classifier trained on real data. Also, we can see points lying outside of the space spanned by real data. They could be meaningful feature vectors that are not in our limited IEMOCAP data. 
\subsubsection{Analysis of 2D projections obtained from $M1$}
\begin{figure*}[t]
\includegraphics[height= 5cm, width=18cm]{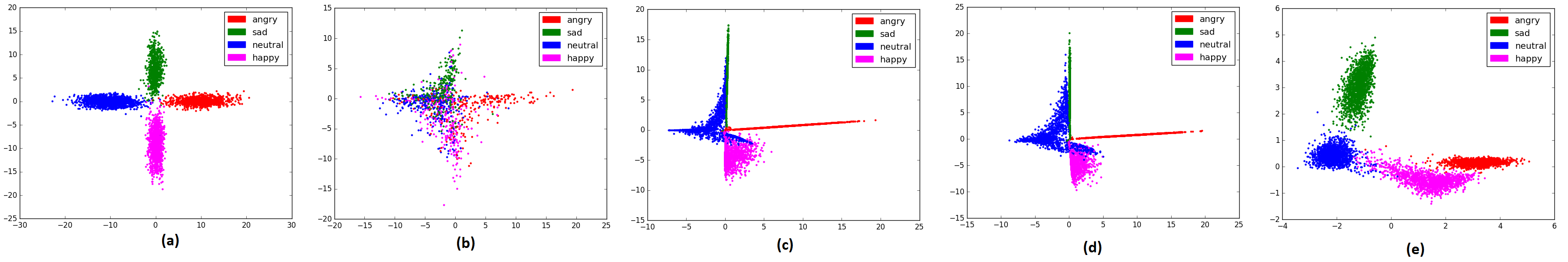}
\caption{Scatter plot of 2D encodings obtained from code space of $M1$ for (a) training set (b) validation set and synthetic data obtained from (c) $M1$ (d) $M2$ (e) $M3$. Note that the scatter plots of synthetic data encodings obtained from $M1$ and $M2$ have a closer resemblance to 2D encodings of training and validation than that of $M3$.}
\label{fig:enc_synth}
\end{figure*}

In this section, we compare the 2D encodings of synthetic data and real data obtained from the trained encoder of $M1$. As mentioned before, $M1$ has been trained so that its code space resembles a mixture of 4-Gaussian. Figure~\ref{fig:enc_synth} shows the scatter plot of 2D encodings obtained for one of the cross-validation splits when the corresponding datasets were passed through the encoder of $M1$. As expected the encodings obtained from the training split strongly resembles a mixture of four Gaussian with each mixture component corresponding to an emotion. The resemblance for encodings obtained from the validation split is not as strong but they are still separable. The scatter plots of encodings obtained from $M1$ and $M2$ look more similar to that of training and validation splits than those obtained from $M3$. It seems as if encodings obtained from $M3$ generated points for a particular emotion lie in clusters which are subsets of the mixture component it is supposed to lie on. However the clusters are farther away from each other with lesser overlap because of the mutual information based loss function trying to maximize inter-class variability. This further indicates that the distributions of synthetic data generated from $M1$ and $M2$ are more similar to real IEMOCAP data than the data generated using $M3$.
\subsection{Cross-corpus experiments}
The objective of cross-corpus evaluations is to investigate the generalization capability added by synthetically generated samples for classification on an external corpus (as opposed to being applicable for only in-domain tasks). To do this we compared the three metrics defined above. We also performed a low resource classification experiment explained below. We generate the synthetic samples from GAN models trained using the entire IEMOCAP dataset. 
\subsubsection{Evaluating the three metrics}
As before, we conduct two experiments (Table~\ref{tab:synth_cc}). First, we use MSP-IMPROV to train an SVM classifier and evaluate it on synthetic data to compute metric 1. Note that since MSP-IMPROV was unbalanced, we balanced it by selecting equal number of audio samples from each class before training the SVM classifier. This was followed with computing metric 2 where we used the synthetic dataset as training set and MSP-IMPROV as test set where we leave it unbalanced. We then computed the FID (metric 3) by comparing the feature vectors obtained from MSP-IMPROV dataset with the synthetic feature vectors. The neural network with similar architecture as described in section was trained with MSP-IMPROV dataset and then the third layer's activations were used to compute FID. 

We observe that evaluating a classifier that has been trained on MSP-IMPROV to classify the synthetic sets shows gives almost similar accuracies for the three models. The slightly higher accuracies obtained for $M2$ and $M3$ could be due to the decoder receiving an extra adversarial error to update its parameters thereby producing more generalizable samples. On the other hand, evaluating different classifiers which has been trained on synthetic samples generated from different GAN based models perform almost similarly in classifying samples from MSP-IMPROV. The FID metric shows a similar trend as the in-domain cross-validation experiment suggesting that synthetic data generated by $M1$ and $M2$ are more similar to the real MSP-IMPROV data than those generated by $M3$. However, the difference between the FID metric obtained from $M3$ generated dataset and the other two models is much lower compared to what we observed in the cross-validation experiment. These experiments indicate that synthetic data generated from the three different GAN architectures trained on a particular dataset compare similarly to data from a different corpus. 
\begin{table}[t]
\centering
\caption {Cross-corpus accuracies obtained on MSP-IMPROV. Synthetic data is generated from GAN based models trained on IEMOCAP}
\vspace{4mm}
\begin{tabular}{@{}l|c|c|c@{}} 
 & Metric 1  & Metric 2 & FID \\ \hline
$M1$      & 48.52 &38.3& 15.91  \\
\hline
$M2$      & 49.6 & 38.61& 15.6 \\
\hline
$M3$     &  49.98& 37.72&18.93 \\
\hline
\end{tabular}
\label{tab:synth_cc}
\end{table}
\subsubsection{Low resource classification experiments}
\begin{figure}[t]
\includegraphics[height=5.5cm, width=9cm]{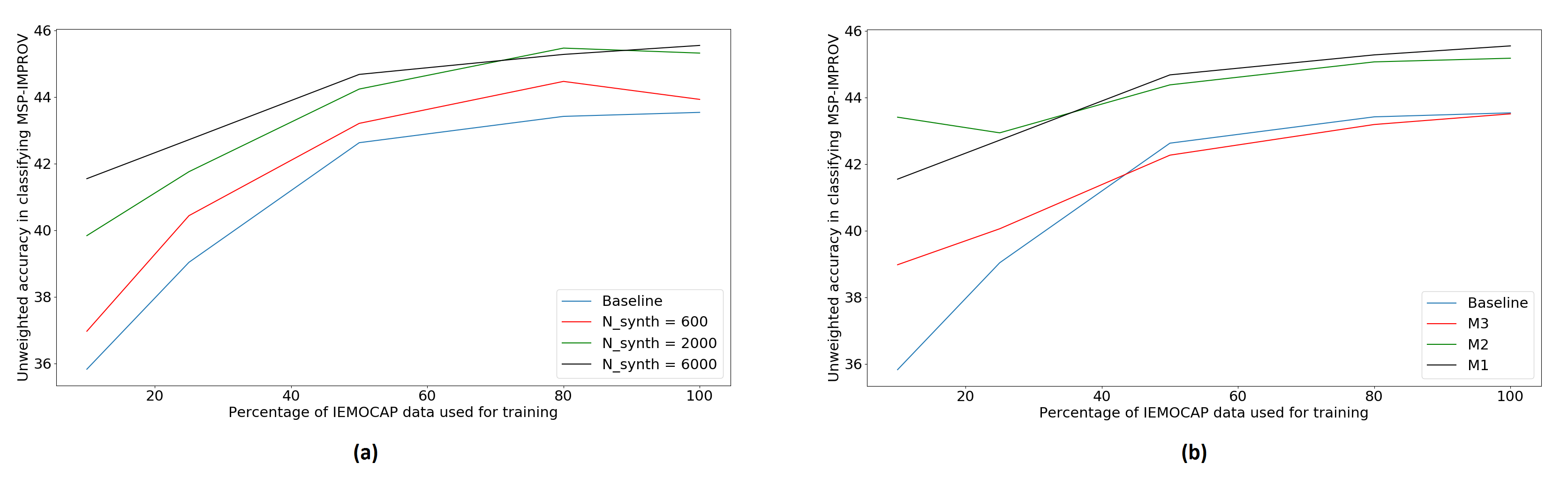}
\caption{Cross-corpus classification accuracy vs the percentage $P$ of real dataset used for training a neural network classifier with or without synthetic data samples obtained from a trained GAN model. Note that the GANs have been trained on the entire IEMOCAP data. (a) Synthetic data from only $M1$ is used while the number of synthetic samples $N_{synth}$ is varied. (b) Synthetic data from all three models $M1$, $M2$ and $M3$ is used with $N_{synth}$ = 6000. Note that for baseline systems $N_{synth} = 0$}
\label{fig:low_1}
\end{figure}

One interesting thing to note is that all the accuracies obtained are higher than the chance accuracy of 25\%. This indicates that synthetic data generated by training a GAN on a specific dataset do carry relevant information which can possibly be leveraged while classifying emotions for unseen data. To validate our hypothesis, we simulated a low resource condition scenario where we use only a portion of IEMOCAP data ($P\%$ of the entire dataset) to train a neural network based classifier and evaluating its performance on MSP-IMPROV. This is our baseline model. Next we append the limited training data with $N_{synth}$ synthetic data samples and repeat the experiment. Figure~\ref{fig:low_1}(a) shows how the accuracies change for different values of $P$ (10\%, 25\%, 50\%, 80\% and 100\%) and $N_{synth}$ (600, 2000 and 6000) when we use synthetic samples generated from $M1$. It can be observed from the figure that using synthetic data along with real data performs better than just using real data for training. Furthermore, the absolute improvement in accuracy is more when lesser amount of real data is used as compared to when the entirety of real data is used for training the neural network based classifier. Note that the synthetic data has been generated from a GAN based model trained on the whole IMEOCAP database. Hence, the synthetic dataset tries to captures the characteristics of the distribution defined by the entirety of the IEMOCAP set. So, it provides more useful information to the classifier while training when only a portion of IEMOCAP is used as opposed to when the whole IEMOCAP dataset is used. We also note that we see more improvement in accuracy when more synthetic samples are used. For lower values of $P$, accuracies keep increasing when we increase $N_{synth}$. For higher values of $P$, they saturate and it seems increasing $N_{synth}$ won't lead to any more improvement in performance. Next we fix $N_{synth}$ at 6000 and compare the performances when we use synthetic data obtained from the three different GAN based models. From Figure~\ref{fig:low_1}(b), we see that the classifiers trained on real data along with samples generated using $M1$ or $M2$ perform similarly while outperforming the baseline. However, samples generated from $M3$ are only beneficial for lower values of P. With availability of larger amount of real data for training the classifier, appending them with synthetic data from $M3$ doesn't lead to any improvement. This can be attributed to the low intra-class variability in the generated samples obtained from $M3$ (as explained in Section~\ref{sec:tsne}) which would lead to the classifier overfitting on only those specific samples present in the training set. Real data with more intra-class variability (as seen in real world) provides more information and lead to a more generalizable classifier. Nevertheless they are still helpful when we have limited real data available for training (P less than 40\% in Figure~\ref{fig:low_1}(b) ). Therefore, our experiments simulating low resource conditions has shown that synthetic data do carry relevant information and can be used for training classifiers when real training data is available in a limited quantity.
\section{Summary and future directions}
\label{sec:conclusion}
In this paper we implemented three auto-encoder and GAN based models to synthetically generate the high-dimensional feature vectors useful for speech emotion recognition. The models were trained to generate such data-points given a sample from a prior distribution $p_{\mathbf{z}}$ . We considered generating synthetic samples for four emotion classes namely angry, sad, neutral and happy. We explored two ways of enforcing a 4-way clustering of the generated data : (a) In models $M1$ and $M2$ where $p_{\mathbf{z}}$ was chosen to be a mixture of four Gaussians with each mixture component corresponding to an emotion class (b) Model $M3$ where $p_{\mathbf{z}}$ was Gaussian but the generator received an additional label vector as input. Mutual information was then maximized between the generated sample and the label vector. In our cross-validation experiments, the FID metric and the experiments classifying synthetic data using SVM trained on real data showed that the distribution $p(\mathbf{x_{synth}},y_{synth})$ generated using $M1$ and $M2$ are closer to the real distribution $p(\mathbf{x_{real}},y_{real})$ than $M3$. Between $M1$ and $M2$, the latter seemed to produce more realistic samples. This can be attributed to the fact the decoder of $M2$ received an extra update based on GAN based adversarial error where a discriminator was used to distinguish between its output and real samples. However, training an SVM using samples generated from $M3$ did a better job in classifying real data-points than $M1$ and $M2$. This was probably because of the tendency of mutual information based loss function resulting in a GAN that would generate samples with more inter-class varibaility. This leads to an SVM trained to have better/more efficient hyper-planes separating the emotional classes. It would be an interesting experiment to explore models where we can possibly take advantage of both these phenomena. In such an experiment, we can define the prior $p_{\mathbf{z}}$ to be a mixture of four Gaussians along with an additional term in the loss function to maximize the mutual information between the generated samples and the mixture component id from which $\mathbf{z} \sim p_{\mathbf{z}}$ was sampled. We can also play around with the weight assigned to the mutual information based loss term relative to the vanilla GAN loss term. The lower dimensional visualizations show that while most of the points lie in the space spanned by real data, a good number of points lie outside of it. In future, we plan to focus more on these data-points to identify the meaningful samples that are not represented in the limited real dataset (IEMOCAP) used to train the GAN based models. The cross-corpus experiments further pointed out that such meaningful samples might exist after all. While $M3$ were only useful when less than 40\% of IEMOCAP data was used for training; samples generated from $M1$ and $M2$ were useful even when all of IEMOCAP data is used for training. This leads us to believe that such GAN based models even though trained with limited real data have the ability to produce meaningful samples which aren't present in the real dataset thereby helping us in cross-corpus emotion recognition. One thing to keep in mind is that more synthetic samples isn't always better as seen from Figure~\ref{fig:low_1}(a) (there is not much difference in accuracies obtained when 2000 and 6000 syntheitc datapoints are used to train a classifier along with the real data.)
Additionally, these synthetic feature vectors cannot be converted back to audio waveforms. Hence we plan to investigate the utility of such architectures to generate audio waveforms corresponding to different emotions. Such samples can be evaluated by having humans listen to them giving us further insights as to how these models behave.


%

\ifCLASSOPTIONcaptionsoff
  \newpage
\fi



%
\bibliographystyle{IEEEbib}
\bibliography{strings}



%
\begin{IEEEbiography}[{\includegraphics[width=0.9in,height=1.25in,clip,keepaspectratio]{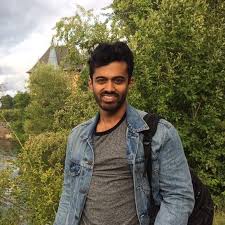}}]{Saurabh Sahu}
received B.Tech and M.Tech degrees in Electronics and Electrical Communication Engineering from Indian Institute of Technology, Kharagpur in 2011 and a Ph.D. degree in Electrical Engineering from the University of Maryland, College Park in 2019. He is currently employed at Samsung Research America. His research interests are in multi-modal sentiment analysis and machine learning. \end{IEEEbiography}

\begin{IEEEbiography}[{\includegraphics[width=0.9in,height=1.25in,clip,keepaspectratio]{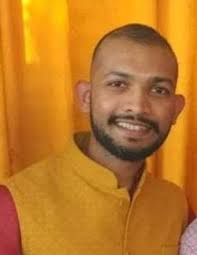}}]{Rahul Gupta}
received a B.Tech. degree in Elec- trical Engineering from Indian Institute of Technology, Kharagpur in 2010 and a Ph.D. degree in Electrical Engineering from University of Southern California (USC), Los Angeles in 2016. His is currently employed at Amazon Alexa and his research concerns development of machine learning algorithms with application to natural language processing. His dissertation work is on the development of computational methods for mod- eling non-verbal communication in human interaction. He is the recipient of Info-USA exchange
scholarship (2009), Provost fellowship (2010-2014) and the Phi Beta Kappa alumni in Southern California scholarship (2015). He was part of the team that won the INTERSPEECH-2013 and INTERSPEECH-2015 Computational Paralinguistics Challenges. He is a member of the IEEE. \end{IEEEbiography}

\begin{IEEEbiography}[{\includegraphics[width=0.9in,height=1.25in,clip,keepaspectratio]{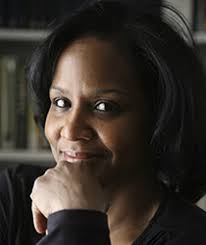}}]{Carol Y. Espy-Wilson}
received a B.S. degree in electrical engineering from Stanford University and M.S., E.E., and Ph.D. degrees in electrical engineering from the Massachusetts Institute of Technology. She is a Professor in the Department of Electrical and Computer Engineering and the Institute for Systems Research at the University of Maryland, College Park. She is also afﬁliated with the Center for Comparative and Evolutionary Biology of Hearing. Her research focuses on understanding the relationship between acoustics and articulation and it involves modeling speech production, studying speech perception, developing signal processing techniques that capture relevant information in speech and using the knowledge gained to develop speech technologies. Current projects include emotion and sentiment analysis, detection and monitoring of depression from speech, and single-channel speech segregation.  Prof. Espy-Wilson is a past Chair of the Speech Technical Committee of the Acoustical Society of America and she is an Associate Editor of the Journal of the Acoustical Society of America.  She is also a past member of the IEEE Speech and Language Technical Committee and she has served as a member of the Language and Communication study section of the National Institutes of Health.   Prof. Espy-Wilson has received several awards including a Clare Boothe Luce Professorship, an NIH Independent Scientist Award, a Distinguished Scholar-Teacher Award from UMD, and a Harvard University Radcliffe Fellowship. She is a Fellow of the Acoustical Society of America and the International Speech Communication Association. 

 \end{IEEEbiography}








\end{document}